\documentclass[twocolumn]{article}
\usepackage{graphicx} 
\usepackage{authblk}
\usepackage{caption}
\usepackage{booktabs}
\usepackage{multirow}
\usepackage{longtable}
\usepackage{geometry}
\usepackage{lscape}
\usepackage{hyperref}
\usepackage[natbib=true,style=numeric-comp,backend=bibtex,useprefix=true]{biblatex}
\title{Dance Dance ConvLSTM}
\author{Miguel O'Malley\thanks{Max Planck Institute for Mathematics in the Sciences, ScaDS.AI Institute of Universitat Leipzig, \texttt{miguel.omalley@mis.mpg.de}}}

\begin{document}

\maketitle

\begin{abstract}
    \textit{Dance Dance Revolution} is a rhythm game consisting of songs and accompanying choreography, referred to as charts. Players press arrows on a device referred to as a dance pad in time with steps determined by the song's chart. In 2017, the authors of Dance Dance Convolution (DDC) developed an algorithm for the automatic generation of \textit{Dance Dance Revolution} charts, utilizing a CNN-LSTM architecture. We introduce Dance Dance ConvLSTM (DDCL), a new method for the automatic generation of DDR charts using a ConvLSTM based model, which improves upon the DDC methodology and substantially increases the accuracy of chart generation.
\end{abstract}

\section{Introduction}
\textit{Dance Dance Revolution} (DDR), and its modern counterpart \textit{In the Groove} (ITG), are rhythm games consisting of songs and charts. Players hit one of four arrows (up, down, left, right) as directed by a chart written to be played in time with an accompanying musical track. Player performance is determined by step accuracy, with steps closer to the assigned timing receiving better evaluations. A single chart may have multiple variations, referred to as difficulty settings, and are associated with a coarser verbal difficulty level (Beginner, Easy, Medium, Hard, Expert) as well as a finer integer difficulty assignment. While the verbal difficulties are generally expected to correllate with increasing difficulty (e.g, Hard should be more challenging than Easy), they do not necessarily correlate across different charts (that is, the Medium difficulty in one chart may be harder than the Challenge difficulty in another.) The fine integer difficulty is expected to be objective; in general songs rated '10' should be harder than songs rated '8', for instance.

\begin{figure}[ht]
    \includegraphics[width = \columnwidth]{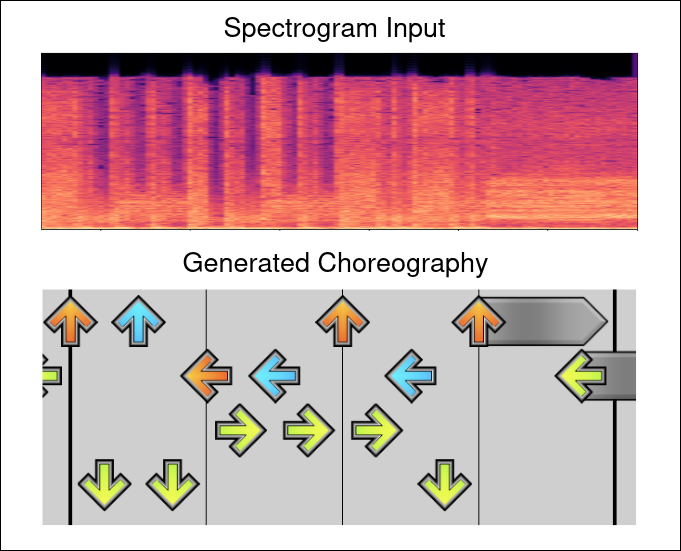}
    \caption{The proposed pipeline takes as input spectrogram information, as above, and returns step placements and selections as below.}
\end{figure}

Charts are expected to correspond to music in a rhythmically fitting manner. In general, steps should land in places where musical context suggests higher activity, while more steps should be placed on higher difficulty levels overall. More complicated (technical) steps should generally be reserved for more eccentric portions of the song, and should not be so densely placed as to render the chart unplayable. We believe these demands are sufficient to constitute a rich task for the application of machine learning (ML), augmented by the extensive breadth and depth of charts publicly available.

The authors of Dance Dance Convolution~\cite{DDC} introduce two models allowing for the automation of the step placement and selection processes, respectively. The DDC step placement model consists of a convolutional encoder taking as input 100ms of audio history and a two layer LSTM decoder. The DDC step selection model consists of a two step LSTM decoder taking as input unrolled previously placed steps, along with basic $\Delta$-time or $\Delta$-beat based information. No audio context is included in the published version of the DDC step selection model, though the DDC authors observed overfitting when convolutional features were included.

DDCL represents a signifcant departure from the DDC methodology. First, we adjust the sampling rate of the input audio context to be $\Delta$-beat based instead of $\Delta$-time. This means that we take samples across each beat, rather than consistently feeding the model 100ms of input information. We achieve this through the application of a traditional onset detection algorithm for the determination of BPM. This adaptation allows us to generate charts for various BPMs, whereas DDC only generates charts at a constant 120BPM. For slower songs, our model takes fewer observations as input, while for faster songs our model will use more observations per second. We further introduce a branched bidirectional ConvLSTM~\cite{ConvLSTM} encoder, allowing us to pass beat-position relevant information in a structurally respectful manner during the encoding process. This allows our model to better track audio contextual information relative to positions of interest within each beat. Audio cues which occur on the downbeat are directly related to the cues which happen on the next downbeat by the ConvLSTM structure. The DDC model outputs thresholded peak based onset predictions based on the music, convolved with a hamming window to suppress double placements and with a 20ms tolerance window for the acccuracy of step placements. Our model produces 48 step placement predictions, spaced evenly across each beat and non-inclusive of the right-hand endpoint. We do not require the usage of a hamming or tolerance window, as the implementation of the BPM detection algorithm together with the full-beat generation architecture allows our model to accurately place steps on the downbeat, offbeat, 16th notes, and so on.

We evaluate our model's performance on the same Fraxtil dataset used in the original version of the DDC paper~\cite{DDC}. Our performance is markedly better in the task of step placement, and also improves significantly in the process of step selection. While our departure from DDC is more considerable in the case of step selection, we consider both cases to represent process improvements through the introduction of ConvLSTM in a music information retrieval (MIR) task. We think this represents potential for a broad advantage held by ConvLSTM in other MIR tasks, since as noted by the authors of DDC~\cite{DDC}, the task of MIR is structurally similar to the task of stepcharting. Through the integration of beat position in the model pipeline, we think ConvLSTM can potentially extract contextually aware information which would otherwise escape or need to be learned silently by $\Delta$-time based models.

\section{Related Work}

Aside from clear predecessor work in DDC~\cite{DDC}, similar approaches have been undertaken for the game Taiko no Tatsujin (TaikoNation~\cite{TaikoNation}), OSU!~\cite{OSU}, BeatMania (GenerationMania~\cite{BeatMania}), and LoveLive~\cite{LoveLive}. Further, an approach extending DDC to generate lower difficulties can be found in DDG~\cite{DDG}, however this approach adapts the DDC methodology and maintains the same generation approach for the top level. Dancing Monkeys~\cite{DancingMonkeys} provides a purely rule based approach to chart generation.

To our knowledge, none of these approaches integrate or examine ConvLSTM as an encoder, or fully resolve the issue of beat placement. DDG addresses the integration of beat-phase through an additional input to the step placement process, but preserves the remainder of the DDC approach.

\section{Data}

We utilize the same Fraxtil dataset from the original DDC~\cite{DDC} paper, consisting of 95 charts from the single eponymous author. Fraxtil's work in these packs is significantly more recent (2013~2014) than the ITG dataset (2004~2005), and thus many of the patterning conventions practiced by Fraxtil are similar to those used presently. We would further argue this presents in the better performance on the Fraxtil dataset observed by the authors of DDC, since the Fraxtil dataset was created in a more structured environment.

We maintain the up-down left-right mirroring process used in the DDC paper. In effect, this process multiplies the volume of step data produced by a given collection of charts by 4. The technicality of a step is unchanged by mirroring all steps left to right or up to down, as patterns which are technical are maintained by this transformation~\cite{ITGwiki}.

We represent our data in two ways for our two distinct models. For the step selection process, we represent each chart as a collection of beats, with each beat represented by the tuple (audio features, placements, auxiliary features). The audio features term consists of a (32, 80, 3) sample of 80 melbands at 32 evenly placed timesteps across the beat. The placements term consists of a variable length string of length at most 48 representing the true step placements taken from the chart across the beat (e.g., 0101 represents beats placed on the 16th note steps but not the quarter or eighth notes.) The auxiliary features term contains two constants relevant to the charting process, (BPM, integer difficulty). This information is critical, as at higher difficulties with lower BPM, steps will be placed more densely. Conversely, at higher BPMs but lower difficulties, step placements will be expected to be more sparse.

\begin{figure}[ht]
    \includegraphics[width = \columnwidth]{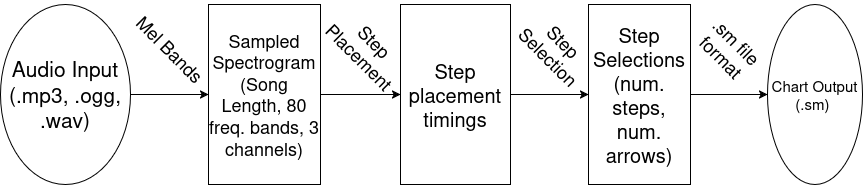}
    \caption{The original DDC architecture. No audio information is used in the selection of steps, and the step placement process is BPM agnostic.}
\end{figure}

\begin{figure}[ht]
    \includegraphics[width = \columnwidth]{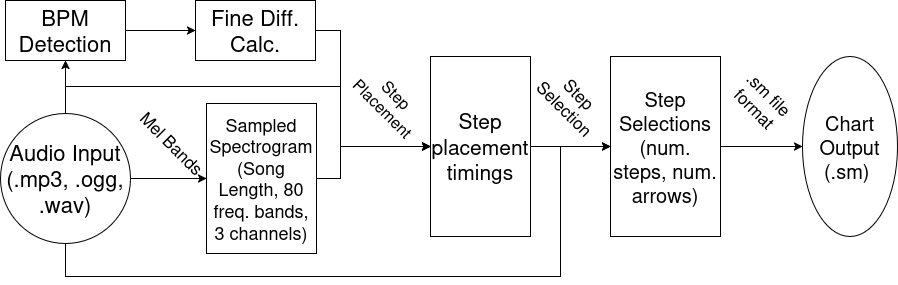}
    \caption{The proposed DDCL architecture. BPM is calculated before the process of step placement through traditional onset detection methods. Difficulty is calculated per chart based on BPM and song length. Audio input is used in the process of step selection.}
\end{figure}

\section{Model description}

Following the DDC methodology, we construct a two part process for our chart generation pipeline. As a pre-generation alignment step, we compute BPM for input songs via the BPM detection algorithm described by van de Wetering~\cite{ArrowVortexPaper} and applied in the charting software ArrowVortex~\cite{ArrowVortex}. ArrowVortex is a commonly used and ubiquitously recommended editor for stepcharting (see~\cite{ITC}). Our step placement model takes as input pure audio features, sampled across each beat. Our goal is to encode each audio feature, using previous and future audio features, and produce steps placed across each beat. Each output is represented as a length 48 binary vector, with 1 representing a placed step and 0 representing an absent step.

Our step selection model takes as input 64 previously placed steps, together with audio features sampled from each step placement and auxiliary features representing the $\Delta$-beat between each step and the following/previous step. For each placed step, there are $4^4=256$ possible steps to choose from. Each direction (left, right, up, down) can either be represented by a 0, 1, 2, or 3, with $0$ representing no step, $1$ representing a placed step, $2$ representing the start of a hold (where the player is expected to keep their foot on the given step until a release appears) and $3$ representing the release from a hold. Many charts have more complicated placements (e.g, ``mines'' which punish the player for stepping on them, and ``rolls'' which require the player to hit them repeatedly until a release step is observed) but in the interest of computational complexity we restrict ourselves to $4$ possibilities for arrow selections.

\section{Methods}

Our generation pipeline is as follows. First, we extract audio features from a given audio file, and determine BPM as described above. We feed these audio features together with BPM and desired difficulty information through our step placement model to determine step placement times. Top default difficulty for any given song is determined by the formula \[ d = \left\lfloor\frac{BPM}{10}\right\rfloor - (4 - \log_2 L)   \] where L is the length of the song. By default, 5 charts will be generated at coarse difficulties Challenge, Hard, Medium, Easy, and Beginner with integer difficulties $d, d-1,\dots,d-4$, respectively. The desired difficulties may also be set by the user. We then pass these step placements together with audio feature information through our step selection model to determine step selections for each chart.

\subsection{Audio features}

In the interest of efficiency, we preserve the audio feature ingestion pipeline introduced in DDC. This process involves ingesting each audio file as a monolithic representation taken as an average of stereo PCM audio. We then maintain three channels for each observation at the input step, representing short-time Fourier transforms (STFT) with windows of 23ms, 46ms, and 92ms, with a stride of 10ms. This methodology mirrors that used in DDC, in turn following methodology from~\cite{BandWindows}. Using Essentia~\cite{Essentia}, we reduce this representation to an 80 band mel-spectrogram representation, with log-scaling to better represent the range of human auditory perception. We then organize this audio information across each beat by taking $32$ evenly spaced samples. For faster songs, it is possible that this can result in duplicated samples due to the sample density. We expand each audio representation in the future and past directions by 15 timesteps, allowing for bidirectional ConvLSTM processing of our audio input.

The final inputs to our step placement model are two tensors of shape (16, 32, 80, 3), representing 16 beat-based timesteps into the future and past, along with two (16, 2) tensors containing the BPM at each beat and the integer difficulty for the chart. We normalize each audio input to have 0-mean and unit standard deviation across each frequency and channel.

\begin{figure}[ht]\label{plot:DDCLstepplacement}
    \includegraphics[width = \columnwidth]{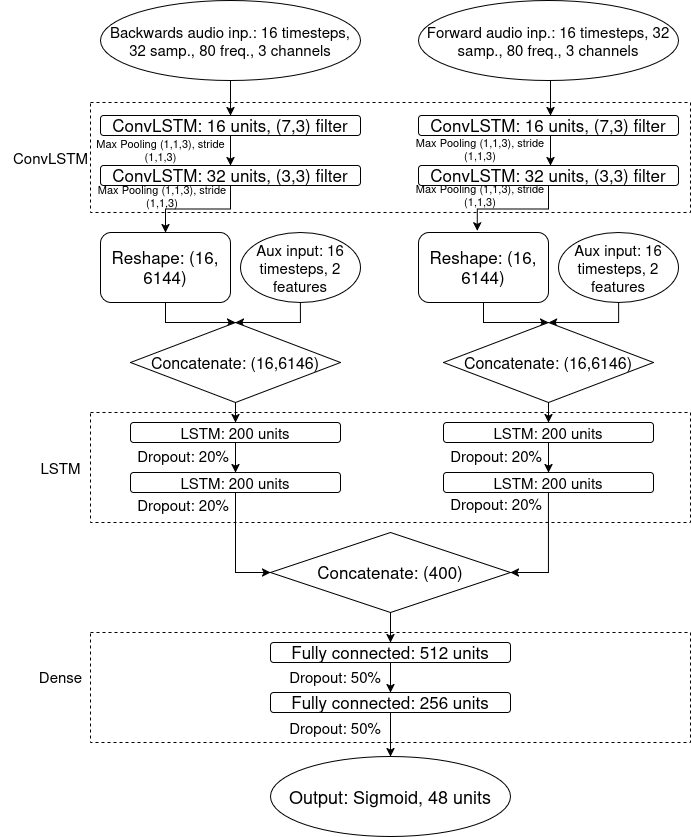}
    \caption{The DDCL step placement model. We employ a branched ConvLSTM encoder structure, along with two LSTM layers and fully connected layers before output. This model places steps one beat at a time, with 48 individual predictions for placements between beats.}
\end{figure}

\subsection{Step placement model}

\subsubsection{Why ConvLSTM?}

Ordinarily, ConvLSTM is used for predictions using video data, as tracking the progress of local features can often provide better foresight than observing all effects at a macro level\cite{ConvLSTM}. It has further been observed in other mediums (see \cite{ConvLSTMcomptime}) that with the same volume of data, ConvLSTM confers significant computational time advantages. Through our construction, ConvLSTM allows for information to be passed in a manner which respects the relative beat position of musical onsets. That is, if a motif repeats on or around the downbeat of a note (such as a drum or bass line), the ConvLSTM structure allows for this information to be directly preserved through the hidden state, without requiring the lower LSTM layers to notice this dependency. Changes in what may otherwise seem subtle things (i.e, percussion, vocal cues, melody) can appear as small changes, or not changes at all if they occur offbeat but at the same volume. All the same, these changes stand out to the listener as significant, due to the shift in an otherwise regular musical sequence. ConvLSTM allows us to better track local changes in motifs, permitting better adherence to the sorts of musical cues which human stepcharters would generally notice and work into charts.

\subsubsection{Full framework}

Our step placement model consists first of two branches of two ConvLSTM layers, of 16 and 32 units respectively, following a similar approach from TaikoNation~\cite{TaikoNation} for a CNN-LSTM model on a similar task. After each layer, we apply 1D maximum pooling in the frequency dimension, with a width and stride of $3$, collapsing similar audio features to focus on larger shifts as the model progresses. We pass the output of our ConvLSTM layers to two branches of two ordinary LSTM layers, after flattening the output of each ConvLSTM layer along all but the timestep dimension and appending the auxiliary input terms along the timestep dimension (BPM, difficulty). We consider this process to be similar to an encoder-decoder structure, where the ConvLSTM layers comprise an encoder for past and future relevant audio information, while the ordinary LSTM layers determine appropriate step placements based on the given difficulty and BPM information as well as the previously encoded audio information. Finally, we concatenate the outputs of both LSTM branches and pass through two fully connected layers with leaky ReLU activation at a negative slope of $0.3$ with 512 and 256 units respectively, then a final output layer of 48 units with sigmoid activation for the final step placement output.

\subsubsection{Training methodology}

We optimize using Adam~\cite{Adam} with an initial learning rate of $.001$ and scheduled reduction on plateau with a factor of $0.5$ and 5 epochs of patience, predicated on the validation set's loss. For loss, we employ binary crossentropy. We train with a batch size of 32 and epoch length of 400 batches. We apply $20\%$ dropout following each normal LSTM layer and $50\%$ dropout following each fully connected layer. Following DDC's methodology, we do not implement any rebalancing scheme to offset the high class imbalance towards null placements. In departure from DDC, we do not explicitly remove beat examples before the first placed step in each chart. Following 100 epochs of warmup, we apply early stopping predicated on the valuation set's precision recall area-under-curve (PR-AUC), stopping after 20 epochs of no improvement and preserving the best performing weights.

\begin{figure}[ht]\label{plot:DDCLstepselection}
    \includegraphics[width = \columnwidth]{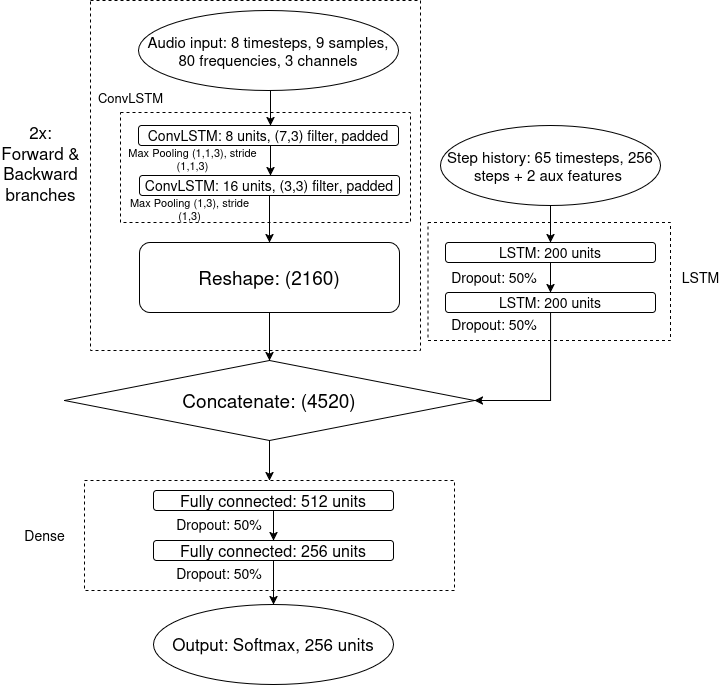}
    \caption{The DDCL step selection model. We employ a branched ConvLSTM encoder structure for audio input, as well as an LSTM for sequential processing using previously placed steps. We concatenate these outputs before feeding to a dense layer for final output.}
\end{figure}

\subsection{Step selection}

Similar to the DDC methodology, we approach the process of step selection as an autoregressive sequence generation procedure, taking the same 64 steps that DDC's model requires. This step history is passed through two LSTM layers consisting of 256 units each. We further augment this procedure with two branches taking as input 8 steps of future and past audio features, which are then passed through two ConvLSTM layers of 8 and 16 units. We flatten and concatenate these outputs, then pass the result through two fully connected layers of 512 and 256 units respectively, with leaky ReLU activation at a negative slope of $0.3$ and 50\% dropout after each fully connected layer. Our output layer consists of a one-hot representation of the 256 possible step combinations, with softmax activation.

\subsection{Training methodology}

Most of our step selection training methodology is identical to that used for the step selection model, aside from a change to a batch size of 64.

\section{Experiments}

We apply the same 8/1/1 split used in the original DDC paper for the Fraxtil dataset. We further maintain the chart difficulty segregation implemented in DDC, as charts for the same song at different difficulty levels often present with similar or identical sections. That is, all difficulties for any given chart will appear in the same split.

\subsection{Step placement evaluation}

We compare our step placement model against two alternative versions intended to identify the particular benefits of design choices made in constructing our model. These versions consist of a Conv3D based encoder instead of a ConvLSTM based encoder, and a fully bidirectional encoder in place of our branched bipartite encoder. We also compare our model to a reconstruction of DDC, with evaluation thresholds at $20$ms and $10$ms. The authors of the original DDC paper have already shown that the DDC methodology outperforms simpler models (based on a logistic regressor, an MLP model, and a pure CNN based model), so we omit comparisons to these approaches.

\begin{figure}[!ht]\label{plot:GenComp}
    \includegraphics[width = \columnwidth]{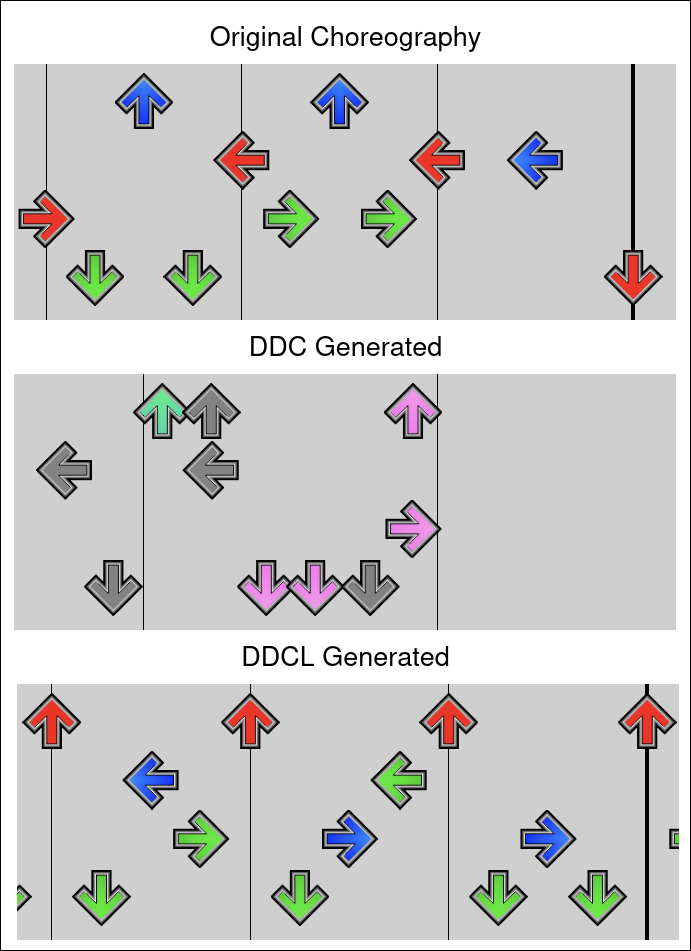}
    \caption{A comparison of DDC and DDCL chart generation outputs for a snippet from the song \textit{Bad Ketchup} by Ladyscraper. The color of each arrow represents the timing placement of the arrow (red represents down beats, blue offbeats, light green 16th notes, etc.) We note the DDC generation exhibits steps placed close together and off-tempo, a central motivation for the structure of DDCL.}
\end{figure}

\subsubsection{Metrics}

We report average performance across all predictions for F1 score, precision, and recall with a threshold set at .5 in Table~\ref{tab:all_metrics}. We also report the same values after selecting the optimal threshold for each chart by metrics with the `max' identifier. Metrics with the `cht' subscript indicate metrics averaged over charts, instead of over total predictions. Over this averaging style, we report precision-recall AUC and loss (binary crossentropy). Additionally, we report model performance averaged across each coarse and fine metric in the validation set in tables~\ref{tab:all_coarse_metrics} and~\ref{tab:all_fine_metrics}, respectively.

We observe that threshold tuning results in the greatest difference for the DDC model's performance. Further, the performance we observe here is somewhat lower than that observed in the original DDC paper for a DDC based model. We suspect this is attributable to differences in validation set composition, as we observe chart-by-chart performance to vary wildly, especially among different difficulties. Especially deleterious for model performance seem to be charts of difficulty $1$ (see table~\ref{tab:all_fine_metrics}), wherein stepcharters place an exceptionally low number of steps.

We further note that the greatest benefit of our $\Delta$-beat based approach seems to be enjoyed amongst the lower difficulty charts. Even with a maximally tuned threshold, the F1 score among charts of difficulty $3$ and $4$ for DDC averages only around $.52$, whereas the F1 score for our ConvLSTM based approach achieves F1 scores averaging around $.8$. Further, little to no benefit is accrued by tuning thresholds at these difficulty levels for our model (see table~\ref{tab:all_fine_metrics}). 

\begin{table}[ht]
    \centering
    \caption{Onset model performance across various metrics.}
    \resizebox{\columnwidth}{!}{
    \begin{tabular}{|l|c|c|c|c|c|}
        \hline
        \textbf{Metric} & \textbf{DDC\textsubscript{(10ms)}} & \textbf{DDC} & \textbf{DDCL} & \textbf{Conv3D} & \textbf{Bidirec.} \\ \hline
        F1Score & 0.5772 & 0.5797 & \textbf{0.7245} & 0.6848 & 0.6518 \\ \hline
        Prec. & 0.6870 & 0.6936 & 0.7439 & 0.7438 & \textbf{0.7529} \\ \hline
        Recall & 0.4976 & 0.4980 & \textbf{0.7060} & 0.6345 & 0.5746 \\ \hline
        Max F1Score & 0.6977 & 0.7127 & \textbf{0.7545} & 0.7524 & 0.7336 \\ \hline
        Max Prec. & 0.6117 & 0.6271 & 0.6665 & \textbf{0.6738} & 0.6551 \\ \hline
        Max Recall & 0.8120 & 0.8255 & \textbf{0.8692} & 0.8517 & 0.8336 \\ \hline
        F1Score\textsubscript{\textbf{cht}} & 0.5200 & 0.5239 & \textbf{0.6266} & 0.6093 & 0.5821 \\ \hline
        Prec\textsubscript{\textbf{cht}} & 0.6157 & 0.6229 & 0.6507 & 0.6796 & \textbf{0.6816} \\ \hline
        Recall\textsubscript{\textbf{cht}} & 0.4975 & 0.5008 & \textbf{0.6264} & 0.5725 & 0.5223 \\ \hline
        Max F1\textsubscript{\textbf{cht}} & 0.6233 & 0.6342 & 0.7073 & \textbf{0.7115} & 0.6877 \\ \hline
        Max Prec.\textsubscript{\textbf{cht}} & 0.5496 & 0.5619 & 0.6244 & \textbf{0.6356} & 0.6122 \\ \hline
        Max Rec.\textsubscript{\textbf{cht}} & 0.7628 & 0.7687 & \textbf{0.8806} & 0.8461 & 0.8412 \\ \hline
        Loss\textsubscript{\textbf{cht}} & 0.0926 & 0.0627 & \textbf{0.0342} & 0.0369 & 0.0402 \\ \hline
        AUC\textsubscript{\textbf{cht}} & 0.5934 & 0.6078 & 0.6749 & \textbf{0.6806} & 0.6607 \\ \hline
        \end{tabular}}
    \caption*{Max modifiers indicate metrics calculated with per-chart thresholds optimized for f1score. Chart averaging indicates metrics averaged across charts instead of across placed steps.}
    \label{tab:all_metrics}
\end{table}

\begin{table}[ht]
    \centering
    \caption{Average sym. model performance by prediction and over charts.}
    \label{tab:sym_avg}
    \resizebox{\columnwidth}{!}{
        \begin{tabular}{@{}c@{}}
            \begin{tabular}{|l|r|r|r|r|r|}
            \hline
            \textbf{Model} & \textbf{Loss} & \textbf{Accuracy} & \textbf{Hold Acc.} & \textbf{Step Acc.} \\
            \hline
            \textbf{DDCL} & \textbf{1.079000} & \textbf{0.581400} & \textbf{0.433800} & \textbf{0.599800} \\
            \hline
            \textbf{DDC} & 1.389600 & 0.497400 & 0.182700 & 0.536900 \\
            \hline
            \textbf{DDCL + Diff.} & 1.098000 & 0.580100 & 0.427900 & 0.599100 \\
            \hline
            \textbf{DDC + CNN} & 1.144100 & 0.558900 & 0.384400 & 0.580800 \\
            \hline
            \end{tabular}
            \\[3em]
            \begin{tabular}{|l|r|r|r|r|r|}
            \hline
            \textbf{Chart Avg.} & \textbf{Loss} & \textbf{Accuracy} & \textbf{Hold Acc.} & \textbf{Step Acc.} \\
            \hline
            \textbf{DDCL} & \textbf{1.008500} & \textbf{0.616000} & \textbf{0.453300} & \textbf{0.628200} \\
            \hline
            \textbf{DDC} & 1.307300  & 0.533400 & 0.210000 & 0.561100 \\
            \hline
            \textbf{DDCL + Diff.} & 1.029400 & 0.611700 & 0.433900 & 0.624600 \\
            \hline
            \textbf{DDC + CNN} & 1.080300  & 0.587600 & 0.391100 & 0.602900 \\
            \hline
            \end{tabular}
        \end{tabular}%
    }
    \caption*{Symbolic model performance averaged across all predictions by prediction and chart. We note the inclusion of difficulty in the symbolic prediction case results in slight overfitting. The inclusion of CNN encoding in the DDC model seems to improve performance substantially, but not to the extent ConvLSTM does.}
\end{table}

\begin{figure}[h]\label{plot:sym_performance}
    \includegraphics[width = \columnwidth]{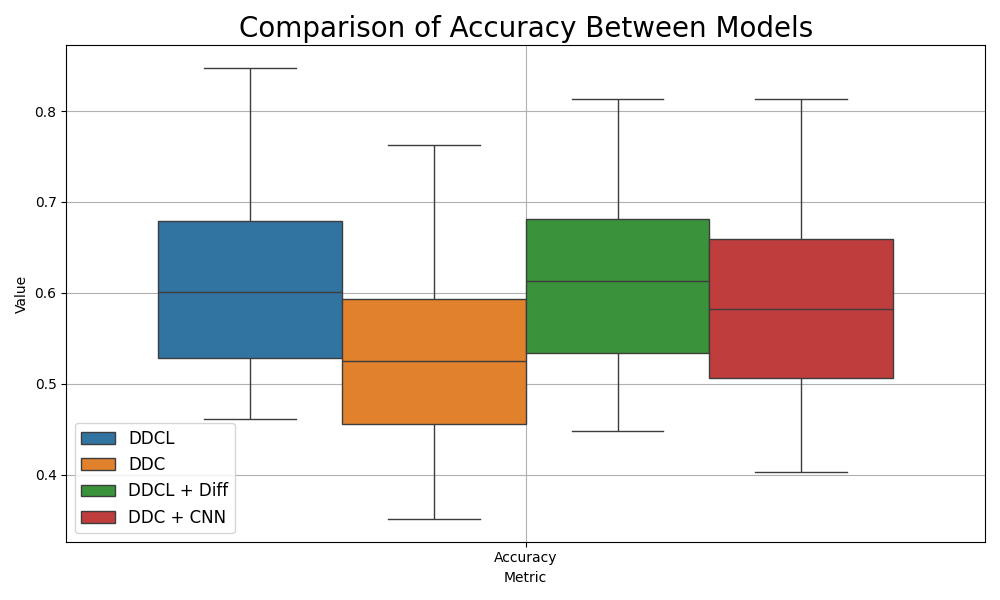}
    \caption{Box plot for the accuracy performance of the evaluated step selection models. We note the distribution is roughly even between models, implying the inclusion of audio features results in uniform improvement in this process.}
\end{figure}
\begin{figure}[h]\label{plot:imp_on_ddc}
    \includegraphics[width = \columnwidth]{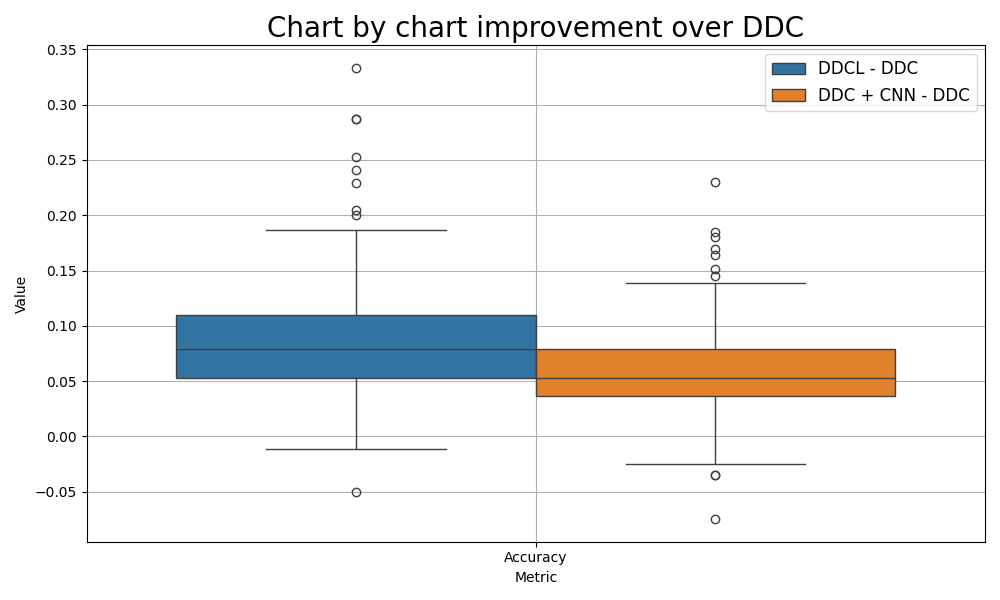}
    \caption{Box plot representing improvement over DDC for the measured approaches. We omit DDCL+Diff, as this model offers little to no improvement over DDCL. We observe that in almost all cases, DDCL improves over DDC, with only the very bottom of the plot representing cases in which DDC has an advantage.}
\end{figure}
\begin{figure}[h]\label{plot:hold_perf}
    \includegraphics[width = \columnwidth]{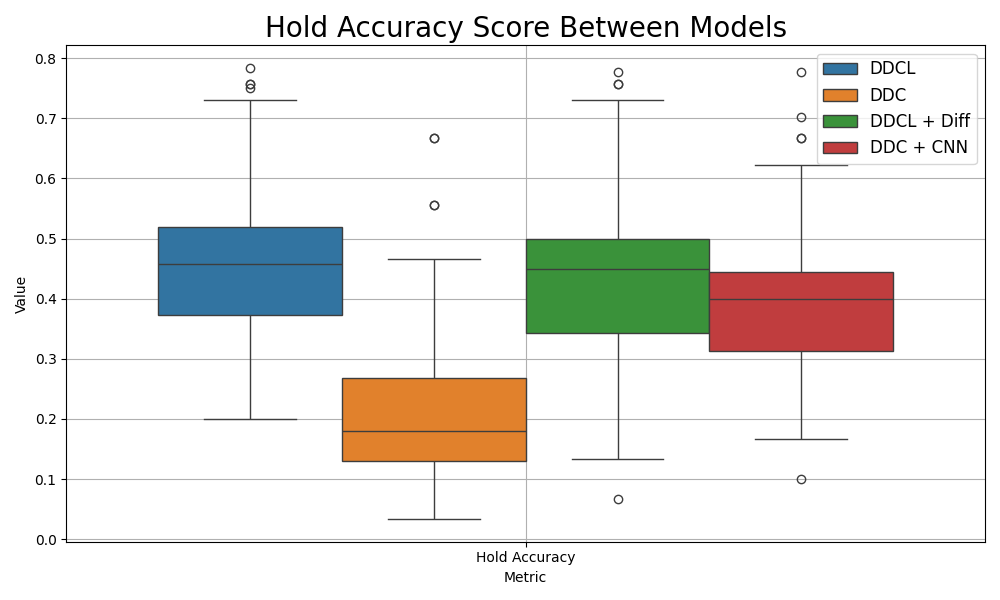}
    \caption{Box plot representing performance on held notes. These less common steps are better represented in the DDCL model, though neither model achieves great performance.}
\end{figure}

\subsection{Step selection evaluation}

For the purposes of evaluation, we compare our model to the version implemented in the DDC paper, as well as a CNN-LSTM based model, and one implementing fine (integer) diffculty as an additional input. The authors of DDC have demonstrated the advantages of the DDC model over simpler approaches to this task (Kneser-Ney smoothed 5-gram models~\cite{KN5}, LSTM with fewer units) so we omit comparisons to these approaches.

\subsubsection{Metrics}

We report average performance across all step selection predictions for accuracy. We further report the distribution of accuracy scores per chart, as well as average accuracy across difficulty levels, and the per chart improvement in accuracy score by model over DDC. We also report accuracy score by step type (i.e, hold vs. step).

We observe that DDCL improves upon DDC across the board. Indeed, DDCL's performance is higher on almost all charts tested. We further observe that concatenating CNN encoded features into the DDC framework before the fully connected layers provides some improvement over DDC, but ConvLSTM provides significantly improved performance over CNN encoding. These performance differentials indictate value both in integrating music features into step selection, and indicate that ConvLSTM is the best method to do so, among those we've tested.

We make note of a significant performance gap between different note types for recorded models. The average accuracy score with which DDC predicts holds (removing charts with no holds at all) is only 20\%. DDCL dramatically improves this score, with an accuracy score of $.45$. While this score is still not particularly good, it is significantly closer to the overall accuracy score. This class difference helps to explain part of the performance difference between DDC and DDCL, but DDCL still improves on non-hold steps as well (though to a lesser degree).

\section{Discussion}

To our knowledge, ConvLSTM has not been used in a pipeline similar to the one we implement here. While there are existing examples of ConvLSTM based audio processing tasks (see \cite{ConvLSTMsoundex1}), we believe the application of ConvLSTM as an encoder for musical audio is novel. The performance improvements displayed by the DDCL architecture suggest that other musical tasks, particularly those requiring the integration of auxiliary inputs (such as difficulty and BPM for DDCL), can benefit significantly from the implementation of a ConvLSTM encoder. It is possible that an encoder like the one we implement here could be used as an input for other music related classification and generation tasks, a topic for future work.

While our model does seem to benefit from threshold tuning, as the original DDC model does, the improvement from doing so is less significant. This is a desirable outcome, as an optimal threshold can only be approximated for generation tasks. Further, the addition of the need for threshold tuning through the use of a validation set is an added hurdle for end users and will vary depending on the training set used. Our performance at a threshold of .5 suggests that in general, a mutable thresholding regime is helpful but not absolutely necessary for decent chart generation performance. Our early-stopping regime mirrors the DDC methodology, though for better threshold tuning it may be desirable to retain models performing best on .5 thresholded F1Score instead of PR-AUC.

The performance of DDCL on step selection tasks improves on the original DDC implementation, but there is likely significant benefit to be gained from training models to handle different charting disciplines. The Fraxtil dataset contains a mixture of technical and stream based charts, which likely creates some confusion in the training process. Future models may benefit from selecting a charting discipline and training a model particular to that task.

While DDCL improves upon the DDC step selection process, it also increases model complexity substantially. The similar DDC+CNN structure we benchmark may be preferable, if the speed of chart generation is a significant concern. Another valid approach could be performing step placement for multiple charts, then processing step selection in ragged batches. The original DDC step selection methodology can also be used along with the DDCL step placement model, though this approach forgoes any convolutional musical encoding.

Our approach still does not represent an end-to-end approach for chart generation, allowing step placement and selection to be handled simultaneously. However, we suspect that such an approach would be both significantly more computationally expensive. Our step placement model is highly parallelizable, whereas the task of step selection (with the current structures) needs to be handled auto-regressively. It is possible that such an approach would increase accuracy, but the two tasks are sufficiently independent for this outcome to be somewhat unlikely.

\section{Conclusions}

The original work in DDC established a pipeline for feasible applications of deep neural netrowks for the purpose of DDR/ITG chart generation. Our modifications to the DDC pipeline resolve sone key issues with the original DDC implementation. Specifically, our results imply the introduction of a ConvLSTM encoder to the DDC pipeline allow for the introduction of fine (integer) difficulty implementation without overfitting. An additional inherent benefit of our process is the integration of variable BPM assignment for different charts. The original DDC implementation only allows for the generation of 120BPM charts, which can feel jarring for players hearing a significantly different tempo. Further, the assignment of steps outside the song's actual tempo results in step placements which occur in palpably odd places in the beat, an issue our framework resolves.

We further demonstrate the value of ConvLSTM as an encoder. ConvLSTM provides performance improvements in both our step placement and selection tasks, suggesting the benefits of ConvLSTM in both onset classification and MIR tasks, both of which resemble our DDCL pipeline.

\section{Code Availability}

All code for this paper can be found at this project's \href{https://github.com/miguelomalley/DDCL}{github}.

\section{Funding}

We gratefully acknowledge partial funding from the Alexander von Humboldt Foundation and the ScaDS.AI institute of Universitat Leipzig.

\section{Acknowledgments}

We express deep gratitude for and commemorate the contributions, support, and memory of Sayan Mukherjee, without whom this work would not have been possible. We thank Alvaro Diaz-Ruelas for helpful comments during the writing of this article. 

\printbibliography
\appendix
\section*{Appendix}

\renewcommand{\thetable}{A\arabic{table}} 
\setcounter{table}{0} 

\onecolumn
\begin{longtable}{|c|c|r|r|r|r|r|}
    \caption{All Metrics by Fine (Integer) Difficulty and Model} \label{tab:all_fine_metrics} \\
    \hline
    \textbf{Metric} & \textbf{Difficulty} & \textbf{DDC\textsubscript{(10ms)}} & \textbf{DDC} & \textbf{DDCL} & \textbf{Conv3D} & \textbf{Bidirec.} \\
    \hline
    \endfirsthead
    
    \hline
    \textbf{Metric} & \textbf{Difficulty} & \textbf{DDC\textsubscript{(10ms)}} & \textbf{DDC} & \textbf{DDCL} & \textbf{Conv3D} & \textbf{Bidirec.} \\
    \hline
    \endhead
    
    \hline
    \multicolumn{7}{r}{\emph{Continued on next page}} \\
    \endfoot
    
    \hline
    \endlastfoot
    \multirow{12}{*}{F1 Score} & 1 & 0.272102 & 0.281437 & 0.145346 & 0.271256 & \textbf{0.290906} \\ 
    & 3 & 0.473902 & 0.484886 & \textbf{0.786356} & 0.693755 & 0.627848 \\ 
    & 4 & 0.444967 & 0.417882 & \textbf{0.784136} & 0.684172 & 0.595607 \\ 
    & 5 & 0.630478 & 0.629469 & \textbf{0.719301} & 0.687575 & 0.625564 \\ 
    & 6 & 0.553643 & 0.545616 & \textbf{0.632485} & 0.600090 & 0.596435 \\ 
    & 7 & 0.563200 & 0.564661 & \textbf{0.730611} & 0.675813 & 0.639810 \\ 
    & 8 & 0.614452 & 0.623302 & \textbf{0.738690} & 0.673271 & 0.666112 \\ 
    & 9 & 0.633272 & 0.638250 & \textbf{0.735156} & 0.681778 & 0.653773 \\ 
    & 10 & 0.651917 & 0.647464 & \textbf{0.713407} & 0.707829 & 0.673422 \\ 
    & 11 & 0.633556 & 0.634376 & \textbf{0.715294} & 0.711280 & 0.678467 \\ 
    & 12 & 0.560409 & 0.591396 & \textbf{0.777388} & 0.732343 & 0.699925 \\ 
    & 13 & 0.624563 & 0.614148 & \textbf{0.782403} & 0.719447 & 0.718592 \\ \hline
   \multirow{12}{*}{Precision} & 1 & 0.189779 & 0.195585 & 0.238689 & \textbf{0.379852} & 0.299484 \\ 
    & 3 & 0.527577 & 0.539613 & 0.702509 & 0.747598 & \textbf{0.756896} \\ 
    & 4 & 0.506792 & 0.458497 & 0.703692 & \textbf{0.706862} & 0.698929 \\ 
    & 5 & 0.769515 & 0.772084 & 0.781281 & 0.791159 & \textbf{0.813724} \\ 
    & 6 & 0.656740 & 0.644136 & 0.642160 & 0.647323 & \textbf{0.670371} \\ 
    & 7 & 0.769467 & 0.777039 & 0.774907 & \textbf{0.792039} & 0.768461 \\ 
    & 8 & 0.701999 & 0.703037 & 0.755214 & 0.736308 & \textbf{0.777225} \\ 
    & 9 & 0.837531 & \textbf{0.863323} & 0.758379 & 0.751934 & 0.778908 \\ 
    & 10 & 0.790844 & 0.792168 & \textbf{0.810745} & 0.732706 & 0.802043 \\ 
    & 11 & 0.819471 & \textbf{0.837073} & 0.745809 & 0.743743 & 0.771220 \\ 
    & 12 & 0.817893 & \textbf{0.865807} & 0.830743 & 0.791114 & 0.836576 \\ 
    & 13 & 0.848079 & \textbf{0.857054} & 0.750700 & 0.800939 & 0.791646 \\ \hline
   \multirow{12}{*}{Recall} & 1 & 0.495999 & \textbf{0.514639} & 0.108099 & 0.255002 & 0.311910 \\ 
    & 3 & 0.440368 & 0.448403 & \textbf{0.902175} & 0.676992 & 0.553689 \\ 
    & 4 & 0.420064 & 0.396857 & \textbf{0.885500} & 0.664320 & 0.525237 \\ 
    & 5 & 0.553060 & 0.549419 & \textbf{0.676186} & 0.614344 & 0.519078 \\ 
    & 6 & 0.478523 & 0.473235 & \textbf{0.623132} & 0.559334 & 0.537323 \\ 
    & 7 & 0.461476 & 0.462173 & \textbf{0.694394} & 0.594743 & 0.556423 \\ 
    & 8 & 0.551402 & 0.563228 & \textbf{0.727959} & 0.623725 & 0.587974 \\ 
    & 9 & 0.518733 & 0.515045 & \textbf{0.723213} & 0.628225 & 0.566211 \\ 
    & 10 & 0.556826 & 0.548929 & 0.652454 & \textbf{0.693293} & 0.595060 \\ 
    & 11 & 0.521876 & 0.515034 & \textbf{0.692688} & 0.683280 & 0.607767 \\ 
    & 12 & 0.428536 & 0.450130 & \textbf{0.732059} & 0.683836 & 0.606661 \\ 
    & 13 & 0.495002 & 0.479093 & \textbf{0.827508} & 0.657739 & 0.660032 \\ \hline
   \multirow{12}{*}{Max F1} & 1 & 0.305789 & 0.316918 & 0.409057 & \textbf{0.474868} & 0.414324 \\ 
    & 3 & 0.563533 & 0.565037 & 0.800973 & \textbf{0.804291} & 0.790504 \\ 
    & 4 & 0.547328 & 0.534026 & 0.801231 & \textbf{0.810609} & 0.778984 \\ 
    & 5 & 0.715432 & 0.720578 & \textbf{0.770029} & 0.765845 & 0.753484 \\ 
    & 6 & 0.632049 & 0.629278 & \textbf{0.696704} & 0.669939 & 0.678940 \\ 
    & 7 & 0.704242 & 0.714304 & \textbf{0.783844} & 0.775117 & 0.735045 \\ 
    & 8 & 0.703881 & 0.718328 & \textbf{0.772199} & 0.752741 & 0.728587 \\ 
    & 9 & 0.790591 & \textbf{0.809423} & 0.784635 & 0.769246 & 0.738012 \\ 
    & 10 & 0.788995 & \textbf{0.793306} & 0.769844 & 0.745988 & 0.741251 \\ 
    & 11 & 0.774197 & \textbf{0.796606} & 0.749091 & 0.740745 & 0.726434 \\ 
    & 12 & 0.753048 & \textbf{0.810000} & 0.802990 & 0.775363 & 0.783777 \\ 
    & 13 & 0.815251 & 0.822598 & \textbf{0.825086} & 0.812041 & 0.821653 \\ \hline
   \multirow{12}{*}{Max Precision} & 1 & 0.209429 & 0.219940 & 0.269778 & \textbf{0.357320} & 0.275803 \\ 
    & 3 & 0.480450 & 0.492807 & 0.695189 & \textbf{0.701383} & 0.687487 \\ 
    & 4 & 0.457870 & 0.450623 & 0.689130 & \textbf{0.695485} & 0.682295 \\ 
    & 5 & 0.674603 & 0.647299 & 0.691611 & \textbf{0.696751} & 0.685893 \\ 
    & 6 & 0.536504 & 0.546087 & 0.555275 & 0.556813 & \textbf{0.587341} \\ 
    & 7 & 0.632339 & 0.650577 & \textbf{0.753705} & 0.730536 & 0.684116 \\ 
    & 8 & 0.620098 & 0.633950 & 0.720340 & \textbf{0.721287} & 0.720320 \\ 
    & 9 & 0.716600 & \textbf{0.749575} & 0.710219 & 0.668844 & 0.650806 \\ 
    & 10 & \textbf{0.725283} & 0.725041 & 0.711840 & 0.706171 & 0.720021 \\ 
    & 11 & 0.723526 & \textbf{0.755594} & 0.718224 & 0.721010 & 0.688846 \\ 
    & 12 & 0.726073 & 0.768755 & \textbf{0.794294} & 0.753261 & 0.760714 \\ 
    & 13 & 0.751215 & \textbf{0.768955} & 0.744522 & 0.710488 & 0.731779 \\ \hline
   \multirow{12}{*}{Max Recall} & 1 & 0.619353 & 0.605171 & \textbf{0.889319} & 0.776859 & 0.849704 \\ 
    & 3 & 0.692200 & 0.704136 & \textbf{0.952105} & 0.947119 & 0.941059 \\ 
    & 4 & 0.773401 & 0.721325 & 0.957876 & \textbf{0.971585} & 0.907964 \\ 
    & 5 & 0.764691 & 0.815503 & \textbf{0.905736} & 0.867346 & 0.856941 \\ 
    & 6 & 0.776899 & 0.758649 & \textbf{0.934928} & 0.843058 & 0.813543 \\ 
    & 7 & 0.797231 & 0.795875 & 0.830629 & \textbf{0.852065} & 0.800128 \\ 
    & 8 & 0.820218 & 0.832853 & \textbf{0.835737} & 0.802159 & 0.751057 \\ 
    & 9 & 0.883141 & 0.880527 & 0.876595 & \textbf{0.905581} & 0.852505 \\ 
    & 10 & 0.867795 & \textbf{0.878862} & 0.845703 & 0.794551 & 0.778144 \\ 
    & 11 & 0.834960 & \textbf{0.843385} & 0.799358 & 0.776489 & 0.774252 \\ 
    & 12 & 0.783901 & \textbf{0.856688} & 0.812029 & 0.800978 & 0.817731 \\ 
    & 13 & 0.891232 & 0.884796 & 0.926204 & \textbf{0.947700} & 0.938355 \\ \hline
    \nopagebreak
   \multirow{12}{*}{Log Loss} & 1 & 0.035210 & 0.034210 & \textbf{0.013128} & 0.013695 & 0.014558 \\ 
    & 3 & 0.034684 & 0.032361 & \textbf{0.013781} & 0.015175 & 0.016748 \\ 
    & 4 & 0.043348 & 0.048147 & \textbf{0.015032} & 0.015968 & 0.017841 \\ 
    & 5 & 0.042210 & 0.037082 & \textbf{0.024831} & 0.028251 & 0.030560 \\ 
    & 6 & 0.066124 & 0.052690 & \textbf{0.029065} & 0.032036 & 0.031895 \\ 
    & 7 & 0.072588 & 0.048244 & \textbf{0.033041} & 0.034346 & 0.040231 \\ 
    & 8 & 0.122834 & 0.094085 & \textbf{0.038287} & 0.044064 & 0.045913 \\ 
    & 9 & 0.114837 & 0.057335 & \textbf{0.046336} & 0.048519 & 0.053871 \\ 
    & 10 & 0.090717 & 0.068899 & \textbf{0.049531} & 0.056888 & 0.058481 \\ 
    & 11 & 0.184005 & 0.100493 & \textbf{0.073492} & 0.075465 & 0.083293 \\ 
    & 12 & 0.375790 & 0.177931 & \textbf{0.078391} & 0.084929 & 0.097649 \\ 
    & 13 & 0.142627 & 0.105576 & \textbf{0.051972} & 0.054716 & 0.060156 \\ \hline
   \multirow{12}{*}{AUC} & 1 & 0.200270 & 0.217418 & 0.258963 & \textbf{0.352673} & 0.277298 \\ 
    & 3 & 0.515131 & 0.511867 & 0.749824 & \textbf{0.757048} & 0.743363 \\ 
    & 4 & 0.458736 & 0.439396 & \textbf{0.725886} & 0.699799 & 0.700765 \\ 
    & 5 & 0.745670 & 0.741401 & 0.778893 & \textbf{0.783668} & 0.776497 \\ 
    & 6 & 0.627178 & 0.619448 & \textbf{0.711747} & 0.658913 & 0.664595 \\ 
    & 7 & 0.696605 & 0.721554 & \textbf{0.775441} & 0.772762 & 0.739696 \\ 
    & 8 & 0.670264 & 0.692466 & \textbf{0.776233} & 0.724147 & 0.723738 \\ 
    & 9 & 0.807476 & \textbf{0.846386} & 0.784357 & 0.772755 & 0.756360 \\ 
    & 10 & 0.780421 & 0.781896 & \textbf{0.809434} & 0.772964 & 0.789164 \\ 
    & 11 & 0.784350 & \textbf{0.813045} & 0.759480 & 0.752585 & 0.747348 \\ 
    & 12 & 0.751402 & 0.830708 & \textbf{0.832213} & 0.805251 & 0.817020 \\ 
    & 13 & 0.849601 & 0.852889 & 0.850267 & \textbf{0.857117} & 0.836596 \\ \hline
\end{longtable}

\begin{longtable}{|c|c|r|r|r|r|r|}
\caption{All Metrics by Coarse Difficulty and Model} \label{tab:all_coarse_metrics} \\
\hline
\textbf{Difficulty} & \textbf{Metric} & \textbf{DDC\textsubscript{(10ms)}} & \textbf{DDC} & \textbf{DDCL} & \textbf{Conv3D} & \textbf{Bidirec.} \\
\hline
\endfirsthead

\hline
\textbf{Difficulty} & \textbf{Metric} & \textbf{DDC\textsubscript{(10ms)}} & \textbf{DDC} & \textbf{DDCL} & \textbf{Conv3D} & \textbf{Bidirec.} \\
\hline
\endhead

\hline
\multicolumn{7}{r}{\emph{Continued on next page}} \\
\endfoot

\hline
\endlastfoot

\multirow{5}{*}{F1 Score} & Beginner & 0.272102 & 0.281437 & 0.145346 & 0.271256 & \textbf{0.290906} \\ 
 & Easy & 0.467472 & 0.469996 & \textbf{0.785863} & 0.691626 & 0.620684 \\ 
 & Medium & 0.590978 & 0.589232 & \textbf{0.703779} & 0.664213 & 0.623840 \\ 
 & Hard & 0.617705 & 0.623465 & \textbf{0.732679} & 0.679238 & 0.660585 \\ 
 & Challenge & 0.627837 & 0.631403 & \textbf{0.740099} & 0.716546 & 0.690568 \\ \hline
\multirow{5}{*}{Precision} & Beginner & 0.189779 & 0.195585 & 0.238689 & \textbf{0.379852} & 0.299484 \\ 
 & Easy & 0.522958 & 0.521587 & 0.702772 & 0.738545 & \textbf{0.744014} \\ 
 & Medium & 0.744438 & 0.745303 & 0.748241 & 0.759489 & \textbf{0.766780} \\ 
 & Hard & 0.743635 & 0.754042 & 0.745519 & 0.733833 & \textbf{0.762943} \\ 
 & Challenge & 0.830155 & \textbf{0.847852} & 0.787791 & 0.766708 & 0.806821 \\ \hline
\multirow{5}{*}{Recall} & Beginner & 0.495999 & \textbf{0.514639} & 0.108099 & 0.255002 & 0.311910 \\ 
 & Easy & 0.435856 & 0.436948 & \textbf{0.898470} & 0.674176 & 0.547366 \\ 
 & Medium & 0.505968 & 0.503407 & \textbf{0.670466} & 0.595586 & 0.535581 \\ 
 & Hard & 0.537318 & 0.540681 & \textbf{0.727298} & 0.638087 & 0.589104 \\ 
 & Challenge & 0.509480 & 0.506868 & \textbf{0.709426} & 0.676359 & 0.608543 \\ \hline
\multirow{5}{*}{Max F1} & Beginner & 0.305789 & 0.316918 & 0.409057 & \textbf{0.474868} & 0.414324 \\ 
 & Easy & 0.559932 & 0.558146 & 0.801031 & \textbf{0.805695} & 0.787944 \\ 
 & Medium & 0.693173 & 0.698198 & \textbf{0.758340} & 0.747623 & 0.730773 \\ 
 & Hard & 0.730639 & 0.745658 & \textbf{0.769229} & 0.753831 & 0.724297 \\ 
 & Challenge & 0.789999 & \textbf{0.812467} & 0.782089 & 0.763951 & 0.764099 \\ \hline
\multirow{5}{*}{Max Precision} & Beginner & 0.209429 & 0.219940 & 0.269778 & \textbf{0.357320} & 0.275803 \\ 
 & Easy & 0.475432 & 0.483433 & 0.693843 & \textbf{0.700072} & 0.686333 \\ 
 & Medium & 0.629826 & 0.625900 & \textbf{0.682012} & 0.676915 & 0.663400 \\ 
 & Hard & 0.649579 & 0.669877 & \textbf{0.705635} & 0.695404 & 0.685253 \\ 
 & Challenge & 0.741103 & \textbf{0.765281} & 0.744174 & 0.727763 & 0.725093 \\ \hline
\multirow{5}{*}{Max Recall} & Beginner & 0.619353 & 0.605171 & \textbf{0.889319} & 0.776859 & 0.849704 \\ 
 & Easy & 0.710244 & 0.707956 & \textbf{0.953388} & 0.952556 & 0.933705 \\ 
 & Medium & 0.778251 & 0.796326 & \textbf{0.887188} & 0.856855 & 0.828359 \\ 
 & Hard & 0.839413 & 0.844662 & \textbf{0.848550} & 0.835962 & 0.781250 \\ 
 & Challenge & 0.847659 & \textbf{0.867306} & 0.834717 & 0.815212 & 0.818129 \\ \hline
\multirow{5}{*}{Log Loss} & Beginner & 0.035210 & 0.034210 & \textbf{0.013128} & 0.013695 & 0.014558 \\ 
 & Easy & 0.036610 & 0.035869 & \textbf{0.014059} & 0.015351 & 0.016991 \\ 
 & Medium & 0.057650 & 0.044271 & \textbf{0.028509} & 0.031124 & 0.034080 \\ 
 & Hard & 0.114500 & 0.078553 & \textbf{0.041773} & 0.045666 & 0.049034 \\ 
 & Challenge & 0.196108 & 0.110148 & \textbf{0.066479} & 0.071106 & 0.077944 \\ \hline
\multirow{5}{*}{AUC} & Beginner & 0.200270 & 0.217418 & 0.258963 & \textbf{0.352673} & 0.277298 \\ 
 & Easy & 0.502599 & 0.495762 & \textbf{0.744505} & 0.744326 & 0.733897 \\ 
 & Medium & 0.702984 & 0.707685 & \textbf{0.762821} & 0.752309 & 0.739363 \\ 
 & Hard & 0.717604 & 0.742708 & \textbf{0.772285} & 0.738071 & 0.729736 \\ 
 & Challenge & 0.798212 & \textbf{0.826839} & 0.806757 & 0.791174 & 0.793783 \\\hline

\end{longtable}

\twocolumn

\end{document}